\documentclass[runningheads]{llncs}
\usepackage{graphicx}
\usepackage{pifont}
\usepackage{indentfirst}
\usepackage{subfigure}
\usepackage{latexsym}

\usepackage{tikz}
\usepackage{comment}
\usepackage{amsmath,amssymb} 
\usepackage{color}
\usepackage{epsfig}
\usepackage{graphicx}
\usepackage{CJK}
\usepackage{bbding}
\usepackage{indentfirst}
\usepackage{multirow}
\usepackage{makecell}
\usepackage{colortbl}
\usepackage{booktabs}

\usepackage{threeparttable}
\usepackage{floatrow}
\usepackage{wrapfig}
\usepackage{lipsum}
\usepackage{caption}
\usepackage{hyperref}

\usepackage{xcolor,colortbl}
\usepackage{MnSymbol,bbding,pifont}
\usepackage[normalem]{ulem}
\useunder{\uline}{\ul}{}

\usepackage[marginal]{footmisc}

\definecolor{Gray}{gray}{0.85}
\definecolor{LightCyan}{rgb}{0.95,0.98,0.98}

\newcolumntype{?}{!{\vrule width 1pt}}

\newcolumntype{a}{>{\columncolor{Gray}}c}
\newcolumntype{b}{>{\columncolor{white}}c}

\usepackage[accsupp]{axessibility}  

\providecommand{\bai}[1]{\textcolor{red}{#1}}

\newcommand{\px}[1]{\textcolor{cyan}{#1}}
\newcommand{\gre}[1]{\textcolor{green}{#1}}
\newcommand{\blu}[1]{\textcolor{blue}{#1}}

\begin{document}

\pagestyle{headings}
\mainmatter
\def\ECCVSubNumber{4252}  

\title{Backbone is All Your Need: A Simplified Architecture for Visual Object Tracking} 
\titlerunning{Backbone is All Your Need}

\author{Boyu Chen\inst{1}$^{, *}$,
Peixia Li\inst{1}$^{, *}$,
Lei Bai\inst{2}$^{, \dagger}$, Lei Qiao\inst{3}, Qiuhong Shen\inst{3}, Bo Li\inst{3}, Weihao Gan\inst{3}, Wei Wu\inst{3}, Wanli Ouyang\inst{2,1}}

\authorrunning{Boyu Chen, Peixia Li et al.}

\institute{The University of Sydney, SenseTime Computer Vision Group, Australia \and
Shanghai AI Laboratory, Shanghai, China \and
SenseTime, China \\ ($^*$) equal contribution; ($^\dagger$) corresponding author\\
\email{bailei@pjlab.org.cn}}
\maketitle

\begin{abstract}
Exploiting a general-purpose neural architecture to replace hand-wired designs or inductive biases has recently drawn extensive interest. However, existing tracking approaches rely on customized sub-modules and need prior knowledge for architecture selection, hindering the development of tracking in a more general system.
This paper presents a \underline{Sim}plified \underline{Track}ing architecture (SimTrack) by leveraging a transformer backbone for joint feature extraction and interaction. Unlike existing Siamese trackers, we serialize the input images and concatenate them directly before the one-branch backbone. Feature interaction in the backbone helps to remove well-designed interaction modules and produce a more efficient and effective framework. To reduce the information loss from down-sampling in vision transformers, we further propose a foveal window strategy, providing more diverse input patches with acceptable computational costs. 
Our SimTrack improves the baseline with 2.5\%/2.6\% AUC gains on LaSOT/TNL2K and gets results competitive with other specialized tracking algorithms without bells and whistles.
The source codes are available at \href{https://github.com/LPXTT/SimTrack}{https://github.com/LPXTT/SimTrack}.

\end{abstract}

\section{Introduction}

Visual Object Tracking (VOT)~\cite{siamban-cvpr20,siamrpnattn-cvpr20,chen2018act,li2018sotsurvey} aims to localize the specified target in a video, which is a fundamental yet challenging task in computer vision.
Siamese network is a representative paradigm in visual object tracking~\cite{Bertinetto-ECCV16-SiamesFC,siamrpn-cvpr18,siamrpnpp,stark}, which usually consists of a Siamese backbone for feature extraction, an interactive head (e.g., naive correlation \cite{Bertinetto-ECCV16-SiamesFC}) for modeling the relationship between the $exemplar$ and \emph{search}, and a predictor for generating the target  localization. 
Recently, transformer~\cite{transt-2021,trans-tem-2021,stark} has been introduced as a more powerful interactive head to Siamese-based trackers for providing information interaction, as shown in Fig.~\ref{fig:m0}(a), and pushes the accuracy to a new level.

\begin{figure*}[t]
\captionsetup{width=1\textwidth}
    \includegraphics[width=0.95\linewidth]{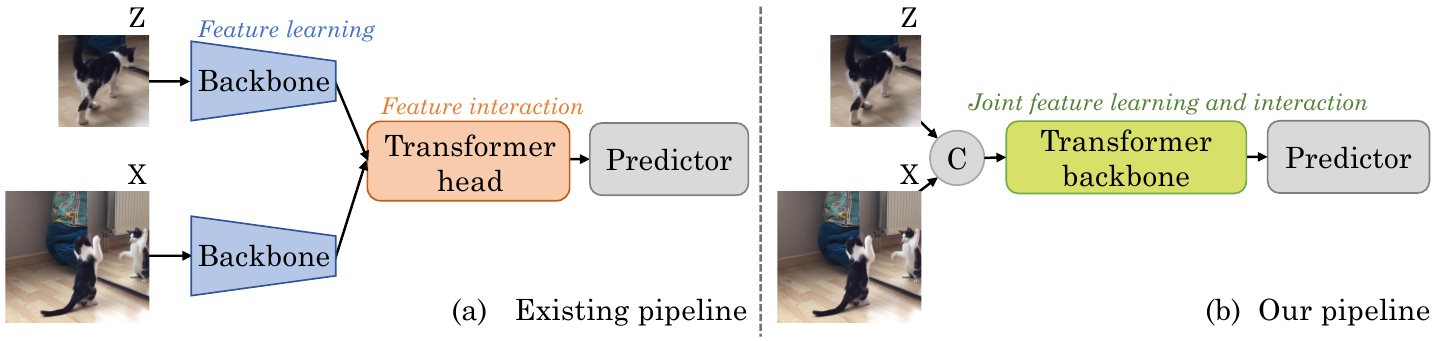}
    \caption{The pipeline of existing transformer trackers (a) and ours (b). A transformer backbone is used to create a simple and generic framework for tracking.}
    \label{fig:m0}
\end{figure*}

While effective, these transformer heads are highly customized and meticulously designed, making it difficult to incorporate them into a more general system or generalize to a wide variety of intelligence tasks.
On the other hand, transformers have recently shown an excellent capability to simplify frameworks for computer vision tasks, like object detection~\cite{chen2021pix2seq} and object segmentation~\cite{zheng2021trans2seg}. Owning to the superior model capacity of transformers, the sub-modules and processes with task-specific prior knowledge can be removed by adequately leveraging transformers to a specific task. Producing a task-agnostic network can not only get a more simplified framework but also help the community move towards a general-purpose neural architecture, which is an appealing trend~\cite{jaegle2021perceiver,zhu2021uniperceiver}. 
However, as observed in this paper, exploiting the transformer to produce a simple and generic framework is not investigated in existing VOT approaches.

With the observation above, this paper advocates a \underline{Sim}plified \underline{Track}ing (SimTrack) paradigm by leveraging a transformer backbone for joint feature learning and interaction, shown as Fig.\ref{fig:m0}(b).
Specifically, we serialize the $exemplar$ ($Z$) and $search$ ($X$) images as multiple tokens at the beginning and send them together to our transformer backbone. Then, the $search$ features from the transformer backbone are directly used for target localization through the predictor without any interaction module. 
Like existing backbones, our transformer backbone can also be pre-trained on other vision tasks, $e.g.$ classification, providing stronger initialization for VOT. Moreover, our SimTrack brings multiple new benefits for visual object tracking. 
(1) Our SimTrack is a simpler and more generic framework with fewer sub-modules and less reliance on prior knowledge about the VOT task. The transformer backbone is a one-branch backbone instead of a Siamese network, consistent with the backbones used in many vision tasks, e.g., image classification~\cite{resnet-cvpr16,vit2020,tang2021mutual,wang2022revisiting}, object detection~\cite{fasterrcnn}, semantic segmentation~\cite{he2017mask,seg2}, depth estimation~\cite{depth1,depth2}, $etc$. 
(2) The attention mechanism in our transformer backbone facilitates a multi-level and more comprehensive interaction between the $exemplar$ and $search$ features. In this way, the backbone features for the $search$ and $exemplar$ image will be dependent on each other in every transformer block, resulting in a designated $examplar$($search$)-sensitive rather than general $search$($examplar$) feature, which is the hidden factor for the effectiveness of the seemingly simple transformer backbone. 
(3) Removing transformer head reduces training expenses. On one hand, the SimTrack can reach the same training loss or testing accuracy with only half training epochs as the baseline model because information interaction happens in a well-initialized transformer backbone instead of a randomly-initialized transformer head.
On the other hand, although adding information interaction in backbone will bring additional computation, the additional computation is generally smaller than that from a transformer head.
(4) According to extensive experiments, SimTrack can get more accurate results with appropriate initialization than other transformer-based trackers using the same transformer as Siamese backbone.

While the transformer-based backbone is capable of achieving sufficient feature learning and interaction between the $exemplar$ and $search$ jointly, the down-sampling operation may cause unavoidable information loss for VOT, which is a localization task and requires more object visual details instead of only abstract$/$semantic visual concepts.
To reduce the adverse effects of down-sampling, we further present a foveal window strategy inspired by fovea centralis.
The fovea centralis is a small central region in the eyes, enabling human eyes to capture more useful information from the central part of vision area.
In our paper, the centre area in the $exemplar$ image contains more target-relevant information and needs more attention accordingly. Therefore, we add a foveal window at the central area to produce more diverse target patches, making the patch sampling frequencies around the image centre higher than those around the image border and improving the tracking performance. 

In conclusion, our contributions are summarized as follows:
\begin{itemize}
	\setlength{\itemsep}{4pt}
	\setlength{\parsep}{4pt}
	\setlength{\parskip}{4pt}
	\item We propose SimTrack, a \underline{Sim}plified \underline{Track}ing architecture that feeds the serialized $exemplar$ and $search$ into a transformer backbone for joint feature learning and interaction. Compared with the existing Siamese tracking architecture, SimTrack only has the one-branch backbone and removes the existing interaction head, leading to a simpler framework with more powerful learning ability. 
	\item We propose a foveal window strategy to remedy the information loss caused by the down-sampling in SimTrack, which helps the transformer backbone capture more details in important $exemplar$ image areas. 
	\item Extensive experiments on multiple datasets show the effectiveness of our method. Our SimTrack achieves state-of-the-art performances with 70.5\% AUC on LaSOT~\cite{lasot-corr}, 55.6\% AUC on TNL2K~\cite{tnl2k-2021}, 83.4\% AUC on TrackingNet~\cite{trackingnet-eccv18}, 69.8\% AO on GOT-10k~\cite{GOT10k} and 71.2\% on UAV123~\cite{uav123-eccv16}.
\end{itemize}

\section{Related Work}
\subsection{Vision Transformer}
Vaswani $et. al.$~\cite{vaswani2017attention} originally proposed transformer and applied it in the machine translation task. The key character of the transformer is the self-attention mechanism which learns the dependencies of all input tokens and captures the global information in sequential data. Thanks to significantly more parallelization and competitive performance, transformer becomes a prevailing architecture in both language modeling~\cite{devlin2018bert,radford2018gpt} and vision community~\cite{vit2020,deit2021,chen2021glit,chen2021psvit}. The first convolution-free vision transformer, ViT~\cite{vit2020}, splits input images into fixed-size patches, which are converted to multiple 1D input tokens. All these tokens are concatenated with a class token and sent into a transformer encoder. After the encoder, the class token is used for image classification. Later, DeiT~\cite{deit2021} introduces a distillation strategy to help transformers reduce the reliance on huge training data. For object detection, DETR~\cite{DETR} treats the task as a sequential prediction problem and achieves promising performance. To reduce the long training time of DETR, deformable DETR~\cite{Defor-Detr} replaces the global attention to adaptive local attention and speeds up the training process. 
Besides, transformer has also shown their powerful potential in other research topics like self-supervised learning~\cite{mocov321,mu2021slip}, multi-module learning~\cite{clip21,kamath2021mdetr}, $etc$.

\subsection{Visual Object Tracking}
Siamese networks is a widely-used two-branch architecture in a surge of tracking algorithms.
Previous works~\cite{Bertinetto-ECCV16-SiamesFC,siamrpn-cvpr18,siamfcpp-aaai20,dasiamrpn-eccv18,gradnet-iccv19,dsiam,siamban-cvpr20,siammask-cvpr20,shen2022unsupervised} based on Siamese Networks~\cite{Bromley-NIPS93-Siamese} formulate VOT as a similarity matching problem and conduct the interaction through cross-correlation. Concretely, SiameseFC~\cite{Bertinetto-ECCV16-SiamesFC} utilize the response map from cross-correlation between the $exemplar$ and $search$ features for target localization. The highest score on the response map generally indicts the target position. In stead of directly getting the target position through the response map, SiamRPN~\cite{siamrpn-cvpr18} and the follow-ups~\cite{dasiamrpn-eccv18,dsiam,siamban-cvpr20,siamrpnattn-cvpr20} send the response map to Region Proposal Network (RPN)~\cite{fasterrcnn} to get a more accurate localization and scale estimation. 
Later, GAT~\cite{gat-cvpr21} and AutoMatch~\cite{nascorr-corr21} tried to replace the global cross-correlation with more effective structure to improve model performance. 
Recently, there have been several notable transformer trackers~\cite{trans-tem-2021,transt-2021,stark} which introduce the transformer to tracking framework for stronger information interaction and achieve compelling results.

All the above-mentioned works introduce interaction between the $exemplar$ and $search$ frames after the backbones. A recent work~\cite{transinfo} adds multiple interaction modellers inside the backbone through hand-designed sub-modules. Our SimTrack also moves information interaction to the backbone but has the following fundamental differences. First, our SimTrack is a more generic and straightforward framework without using Siamese architecture or well-designed interaction modules, which are both used in \cite{transinfo} and all above Siamese-based methods. Second, our SimTrack utilizes pre-trained vision transformers for the interaction instead of training the interaction module from scratch. Third, the interaction between the $exemplar$ and $search$ exists in each block of our backbone. In contrast, the interaction modules are only added at the end of several blocks in \cite{transinfo}. Fourth, there is only information flow from the $exemplar$ feature to the $search$ feature in \cite{transinfo}, while ours has bidirectional information interaction between the $exemplar$ and $search$ features.

\section{Proposed Method}
\begin{figure*}[t]
    \centering
    \includegraphics[width=0.98\linewidth]{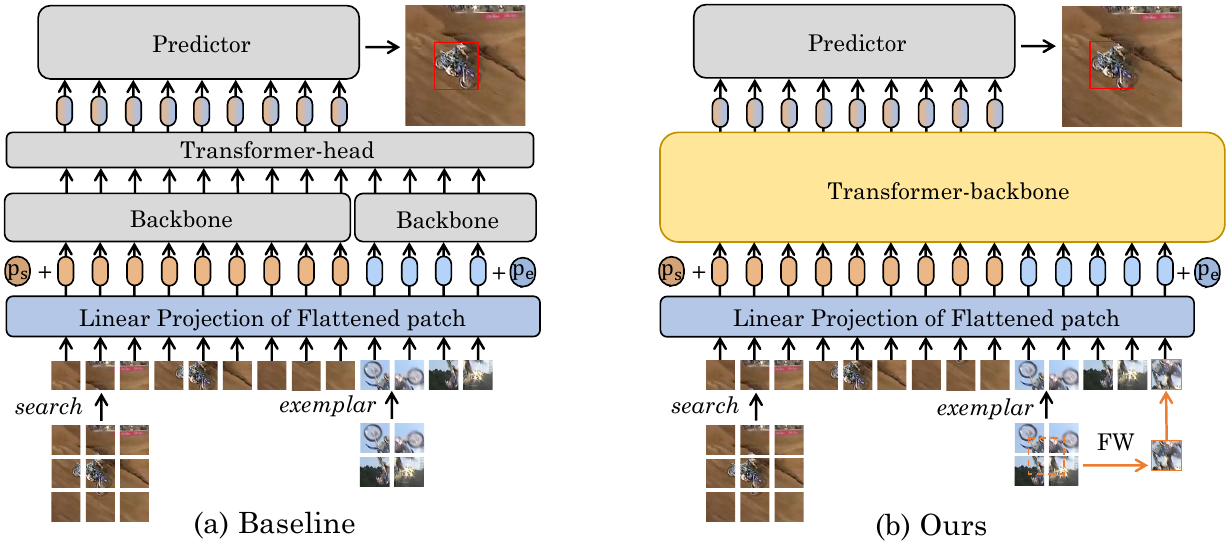}\\
    \caption{The pipeline of the baseline model (a) and our proposed SimTrack (b). `FW' in (b) denotes foveal window, $p_s$ and $p_e$ are position embedding of the $search$ and $exemplar$ tokens. In (b), a transformer backbone is utilized to replace the Siamese backbone and transformer head in (a). Both $exemplar$ and $search$ images in (b) are serialized into input sequences, which are sent to the transformer backbone for joint feature extraction and interaction. Finally, the target-relevant search feature is used for target localization through a predictor.}
    \label{fig:framework}
\end{figure*}

Our SimTrack consists of a transformer backbone and a predictor, as shown in Fig.~\ref{fig:framework} (b). The transformer backbone is used for feature extraction and information interaction between the $exemplar$ and $search$ features, guiding the network to learn a target-relevant $search$ feature. After passing the backbone, the output features corresponding to the $search$ area are sent to a corner predictor for target localization. For better understanding, we will first introduce our baseline model in Sec.~\ref{sub:preliminary}, which replaces the CNN backbone of STARK-S~\cite{stark} with a transformer backbone, and then show details of our SimTrack in Sec.~\ref{sub:simtrack} and the foveal window strategy for improving SimTrack in Sec.~\ref{sub:att}.

\subsection{Baseline Model}
\label{sub:preliminary}
STARK-S has no extra post-processing during inference,
which is consistent with our initial purpose to simplify the tracking framework.
We replace the backbone of STARK-S~\cite{stark} from Res50~\cite{resnet-cvpr16} to ViT~\cite{vit2020} to get our baseline model STARK-SV.
Like other transformer-based trackers, the pipeline of STARK-SV is shown in Fig.~\ref{fig:m0} (a). Given a video, we treat the first frame with ground truth target box as $exemplar$ frame. According to the target box, we crop an $exemplar$ $\mathbf{Z}\in\mathbb{R}^{{H_z}\times{W_z}\times{3}}$ from the first frame, where (${H_z}$, ${W_z}$) is the input resolution of $\mathbf{Z}$. All following frames $\mathbf{X}\in\mathbb{R}^{{H_x}\times{W_x}\times{3}}$ are the $search$ frames.

{\flushleft{\textbf{Image serialization.}}}
The two input images are serialized into input sequences before the backbone. 
Specifically, similar to current vision transformers~\cite{vit2020,deit2021}, we reshape the images $\mathbf{Z}\in\mathbb{R}^{{H_z}\times{W_z}\times3}$ and $\mathbf{X}\in\mathbb{R}^{{H_x}\times{W_x}\times3}$ into two sequences of flattened 2D patches $\mathbf{Z_p}\in\mathbb{R}^{N_z\times({P^2}\cdot{3})}$ and $\mathbf{X_p}\in\mathbb{R}^{N_x\times({P^2}\cdot{3})}$, where $(P,P)$ is the patch resolution, $N_z={H_z}{W_z}/P^2$ and $N_x={H_x}{W_x}/P^2$ are patch number of the $exemplar$ and $search$ images.
The 2D patches are mapped to 1D tokens with $C$ dimensions through a linear projection.
After adding the 1D tokens with positional embedding~\cite{vaswani2017attention}, we get the input sequences of the backbone, including the $exemplar$ sequence $e^0\in\mathbb{R}^{N_z\times{C}}$ and the $search$ sequence $s^0\in\mathbb{R}^{N_x\times{C}}$.

{\flushleft{\textbf{Feature extraction with backbone.}}}
The transformer backbone consists of $L$ layers. We utilize $e^l$ and $s^l$ to represent the input $exemplar$ and $search$ sequences of the $(l+1)_{th}$ layer, $l=0,...,L-1$. The forward process of the $exemplar$ feature in one layer can be written as:
\begin{equation}
\begin{gathered}
    e^* = e^{l} + Att(LN(e^{l})), \\
    e^{l+1} = e^* + FFN(LN(e^*)), 
\end{gathered}
\label{equ:el}
\end{equation}
where $FFN$ is a feed forward network, $LN$ denotes Layernorm and $Att$ is self-attention module~\cite{vaswani2017attention} (we remove $LN$ in the following functions for simplify),
\begin{equation}\label{equ:att}
    Att(e^l) = softmax\left(\frac{{(e^l}W_Q)({{e^l}W_K})^T}{\sqrt{d}}\right)\left({e^l}W_V\right),
\end{equation}
where $1/\sqrt{d}$ is the scaling factor, $W_Q\in\mathbb{R}^{C\times{D}}$, $W_K\in\mathbb{R}^{C\times{D}}$, $W_V\in\mathbb{R}^{C\times{D}}$ are project metrics to convert input sequence to $query$, $key$ and $value$.
Generally, multi-head self-attention~\cite{vaswani2017attention} is adopted to replace self-attention in Equ.(\ref{equ:el}). For simplicity and better understanding, we use the self-attention module in our descriptions.
As we can see, the feature extraction of $e^l$ only considers $exemplar$ information.
The feed forward process of $s^l$ is the same as $e^l$.
After passing the input into the backbone, we get the output $exemplar$ sequence $e^L$ and the output $search$ sequence $s^L$.

{\flushleft{\textbf{Feature interaction with transformer head.}}}
The features $e^L\in\mathbb{R}^{N_z\times{D}}$ and $s^L\in\mathbb{R}^{N_x\times{D}}$ interact with each other in the transformer head. 
We refer readers to STARK-S~\cite{stark} for more details of the transformer head in our baseline models.

{\flushleft{\textbf{Target localization with predictor.}}}
After transformer head, we get a target-relevant $search$ feature $s^{L*}\in\mathbb{R}^{N_x\times{D^*}}$, which is reshaped to ${\frac{H_x}{s}}\times{\frac{W_x}{s}\times{D^*}}$ and sent to a corner predictor. The corner predictor outputs two probability maps for the top-left and bottom-right corners of the target box.

During offline training, a pair of images within a pre-defined frame range in a video are randomly selected to serve as the $exemplar$ and $search$ frame. After getting the predicted box $b_i$ on the $search$ frame, the whole network is trained through $\ell_1$ loss and generalized IoU loss~\cite{DETR},
\begin{equation}\label{eq:train}
L = \lambda_{iou}L_{iou}(b_i, b_i^{*}) + \lambda_{L_1}L_{1}(b_i, b_i^{*}),
\end{equation}
where $b_i^*$ is the ground truth box, $\lambda_{iou}$ and $\lambda_{L_1}$ are loss weights, $L_{iou}$ is generalized IoU loss and $L_{1}$ is the $\ell_1$ loss.

\subsection{Simplified Tracking Framework}
\label{sub:simtrack}
Our key idea is replacing the Siamese backbone and transformer head in the baseline model with a unified transformer backbone, as shown in Fig.~\ref{fig:framework} (b).
For STARK-S, the function of the backbone is to provide a strong feature extraction. The transformer head is responsible for information interaction between the $exemplar$ and $search$ features.
In our SimTrack, only a transformer backbone is needed for joint feature and interaction learning. 
In the following, we show how to apply vision transformer as a powerful backbone to VOT successfully and create a more simplified framework.
The input of our transformer backbone is also a pair of images, the $exemplar$ image $\mathbf{Z}\in\mathbb{R}^{{H_z}\times{W_z}\times3}$ and the $search$ image $\mathbf{X}\in\mathbb{R}^{{H_x}\times{W_x}\times3}$. Similarly, we first serialize the two images to input sequences $e^0\in\mathbb{R}^{N_z\times{C}}$ and $s^0\in\mathbb{R}^{N_x\times{C}}$ as mentioned above.

{\flushleft{\textbf{Joint feature extraction and interaction with transformer backbone.}}}
Different from the baseline model, we directly concatenate $e^0$ and $s^0$ along the first dimension and send them to the transformer backbone together. The feed forward process of $(l+1)_{th}$ layer is:
\begin{equation}
{
\begin{gathered}
\left[ \begin{array}{c} e^* \\ s^* \end{array} \right] = \left[\begin{array}{c} e^l \\ s^l \end{array}\right] + Att\left(\left[ \begin{array}{c} e^l \\ s^l \end{array} \right]\right), \\
\left[ \begin{array}{c} e^{l+1} \\ s^{l+1} \end{array} \right] = \left[\begin{array}{c} e^* \\ s^* \end{array}\right] + FFN\left(\left[ \begin{array}{c} e^* \\ s^* \end{array} \right]\right).
\end{gathered}
\label{equ:ela}
}
\end{equation}
The symbol of layer normalization is removed in Equ.(\ref{equ:ela}) for simplify. The main difference between Equ.(\ref{equ:el}) and Equ.(\ref{equ:ela}) is the computation in $Att(.)$,
\begin{equation}
\begin{gathered}
Att(\left[ \begin{array}{c} e^l \\ s^l \end{array} \right]) = softmax\left(\begin{bmatrix}
a(e^l, e^l), & a(e^l, s^l) \\
a(s^l, e^l), & a(s^l, s^l)
\end{bmatrix}\right) \left(\left[ \begin{array}{c} e^lW_V \\ s^lW_V \end{array} \right]\right),\\
\end{gathered}
\label{equ:atta}
\end{equation}
where $a(x, y) = (xW_Q)(yW_K)^T/{\sqrt{d}}$.
After converting Equ.(\ref{equ:atta}), the $exemplar$ attention $Att(e^l)$ and the $search$ attention $Att(s^l)$ are,
\begin{equation}
\begin{gathered}
Att(e^l) = softmax\left(\left[a(e^l, e^l), a(e^l, s^l)\right]\right)\left[ e^lW_V, s^lW_V \right]^T, \\
Att(s^l) = softmax\left(\left[a(s^l, e^l), a(s^l, s^l)\right]\right)\left[ e^lW_V, s^lW_V \right]^T.
\end{gathered}
\label{equ:attesnew}
\end{equation}

In the baseline model, the feature extraction of the $exemplar$ and $search$ features are independent with each other as shown in Equ.(\ref{equ:att}). While, in our transformer backbone, the feature learning of $exemplar$ and $search$ images influence each other through $a(e^l, s^l)$ and $a(s^l, e^l)$ in Equ.(\ref{equ:attesnew}). $Att(e^l)$ contains information from $s^l$ and vice verse. The information interaction between the $exemplar$ and $search$ features exists in every layer of our transformer backbone, so there is no need to add additional interaction module after the backbone. We directly send the output $search$ feature $s^L$ to the predictor for target localization.

{\flushleft {\bf{Distinguishable position embedding. }}}
It is a general paradigm to seamlessly transfer networks pre-trained from the classification task to provide a stronger initialization for VOT. In our method, we also initialize our transformer backbone with pre-trained parameters.
For the $search$ image, the input size ($224\times224$) is the same with that in general vision transformers~\cite{vit2020,deit2021}, so the pre-trained position embedding $p_0$ can be directly used for the $search$ image ($p_s = p_0$). However, the $exemplar$ image is smaller than the $search$ image, so the pre-trained position embedding can not fit well for the $exemplar$ image. Besides, using the same pre-trained position embedding for both images provides the backbone with no information to distinguish the two images. To solve the issue, we add a learnable position embedding $p_e\in\mathbb{R}^{N_z \times D}$ to the $exemplar$ feature, which is calculated by the spatial position $(i, j)$ of the patch and the ratio $R_{ij}$ of the target area in this patch (as depicted in Fig.~\ref{fig:s} (b)),
\begin{equation}\label{equ:ps}
    p_{e} = FCs(i, j, R_{ij}),
\end{equation}
where $p_e$ denotes the position embedding of the $exemplar$ feature, $FCs$ are two fully connected layers. After obtaining the position embedding $p_e$ and $p_s$, we add them to the embedding vectors. The resulting sequences of embedding vectors serve as inputs to the transformer backbone.

\subsection{Foveal Window Strategy}
\label{sub:att}
\begin{wrapfigure}{r}{0.6\textwidth}
	\centering
	\begin{minipage}[c]{6cm}
		\includegraphics[width=7cm]{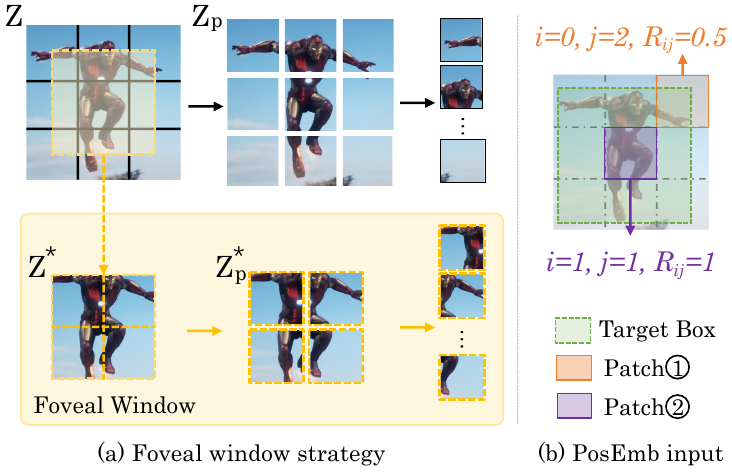}
	\end{minipage}%
	\hfill
	\begin{minipage}[c]{6cm}
        \caption{(a) the foveal window strategy and (b) getting the inputs of $FCs$ in Equ.(\ref{equ:ps}).}\label{fig:s}
	\end{minipage}%
\end{wrapfigure}

The $exemplar$ image contains the target in the center and a small amount of background around the target. The down-sampling process may divide the important target region into different parts.
To provide the transformer backbone with more detailed target information, we further propose a foveal window strategy on the $exemplar$ image to produce more diverse target patches with acceptable computational costs. As shown in the second row of Fig.~\ref{fig:s}(a), 
we crop a smaller region $\mathbf{Z^*}\in\mathbb{R}^{{H_z}^*\times{W_z}^*\times3}$ in the center of the $exemplar$ image and serialize $\mathbf{Z^*}$ into image patches $\mathbf{Z_p^*}\in\mathbb{R}^{N_z^*\times({P^2}\cdot{3})}$, where $N_x^*={H_x^*}{W_x^*}/P^2$. 
The partitioning lines on $\mathbf{Z^*}$ are located in the center of those on the $exemplar$ image $\mathbf{Z}$, so as to ensure that the foveal patches $\mathbf{Z_p^*}$ contain different target information with the original patches $\mathbf{Z_p}$. 
After getting the foveal patches $\mathbf{Z_p^*}$, we calculate their position embedding according to Equ.(\ref{equ:ps}). Then, we map $\mathbf{Z_p^*}$ with the same linear projection as $\mathbf{Z_p}$ and add the mapped feature with the position embedding to get the foveal sequence $e^{0*}$. Finally, the input of transformer backbone includes the $search$ sequence $s^0$, the $exemplar$ sequence $e^0$ and the foveal sequence $e^{0*}$.
The $exemplar$ image is small in VOT, so the token number in $e^0$ and $e^{0*}$ are modest as well.

\section{Experiments}

\subsection{Implementation Details}
\textbf{Model.}
We evaluate our method on vision transformer~\cite{clip21} and produce three variants of SimTrack: Sim-B$/$32, Sim-B$/$16, and Sim-L$/$14 with the ViT base, base, and large model~\cite{vit2020} as the backbone, respectively, where input images are split into $32\times{32}$, $16\times{16}$ and $14\times{14}$ patches, correspondingly. All parameters in the backbone are initialized with pre-trained parameters from the vision branch of CLIP~\cite{clip21}.
For better comparison with other trackers, we add another variant Sim-B$/$16$^*$ with fewer FLOPs than Sim-B$/$16. In Sim-B$/$16$^*$, we remove the last four layers in the transformer backbone to reduce computation costs.
The predictor is exactly the same as that in STARK-S~\cite{stark}.

\textbf{Training.}
Our SimTrack is implemented with Python 3.6.9 on PyTorch 1.8.1. All experiments are conducted on a server with 8 16GB V100 GPUs. The same as STARK-S, we train our models with training-splits of LaSOT~\cite{lasot-corr}, GOT-10K~\cite{GOT10k}, COCO2017~\cite{coco}, and TrackingNet~\cite{trackingnet-eccv18} for experiments on all testing datasets except for GOT-10k\_Test. For GOT-10k\_Test, we follow the official requirements and only use the $train$ set of GOT-10k for model training.
In Sim-B$/$32, we set the input sizes of $exemplar$ and $search$ images as $128\times128$ and $320\times320$, corresponding to $2^2$ and $5^2$ times of the target bounding box, because the larger stride 32 makes the output features having a smaller size. Too small output size has a negative effect on target localization.
In Sim-B$/$16, the input sizes are $112\times112$ and $224\times224$, corresponding to $2^2$ and $4^2$ times of the target bounding box.
For Sim-L$/$14, the $exemplar$ input size is reduced to $84\times84$ ($1.5^2$ times of target bounding box) to reduce computation costs. 
Without the special declaration, all other experiments use the same input sizes as Sim-B$/$16.
 The size of the cropped image for the foveal window is $64\times64$.
 All other training details are the same with STARK-S~\cite{stark} and shown in the supplementary materials.

\textbf{Inference.}
Like STARK-S~\cite{stark}, there is no extra post-processing for all SimTrack models. The inference pipeline only consists of a forward pass and coordinate transformation process. The input sizes of $exemplar$ and $search$ images are consistent with those during offline training. Our Sim-B/16 can run in real-time at more than 40 $fps$.

\subsection{State-of-the-art Comparisons}
\begin{table*}[t]
\begin{center}
\caption{Performance comparisons with state-of-the-art trackers on the $test$ set of LaSOT~\cite{lasot-corr}, TNL2K~\cite{tnl2k-2021} and TrackingNet~\cite{trackingnet-eccv18}. `Size' means the size of $search$ image, `FLOPs' shows the computation costs of backbone and transformer head. For methods without transformer head, `FLOPs' shows the computation costs from the backbone. AUC, P$_{norm}$ and P are AUC, normalized precision and precision. Sim-B$/16^*$ denotes removing the last four layers of the transformer-backbone in Sim-B$/16$ to reduce FLOPs. Trackers shown with $\Diamond$ have online update modules. \bai{Red}, \gre{green} and \blu{blue} fonts indicate the top-3 methods.}
\label{table:sota}
\scalebox{0.90}{
\setlength{\tabcolsep}{1.5mm}{
\begin{tabular}{l|c|c|r|c|c|c|c|c|c}
\Xhline{1pt}
\multirow{2}*{Methods}  & \multirow{2}*{Net}  & \multirow{2}*{Size} & \multirow{2}*{FLOPs} & \multicolumn{2}{c|}{LaSOT} & \multicolumn{2}{c|}{TNL2K} & \multicolumn{2}{c}{TrackingNet} \\ \cline{5-10}
&  & & & AUC & P$_{norm}$ & AUC & P & AUC & P\\   
\hline
SiamFC~\cite{Bertinetto-ECCV16-SiamesFC}  & AlexNet  & 255 & 4.9G & 33.6 & 42.0 & 29.5 & 28.6 &  57.1&  66.3\\
ATOM~\cite{atom-cvpr19} $\Diamond$  & ResNet18  & 288 & 3.0G & 51.5 & 57.6 & 40.1 & 39.2 & 70.3 & 64.8\\
DiMP~\cite{dimp-iccv19} $\Diamond$  & ResNet50  & 288 & 5.4G & 56.9 & 65.0 & 44.7 & 43.4 & 74.0 & 68.7\\
SiamRPN$++$~\cite{siamrpnpp}  & ResNet50  & 255 & 7.8G & 49.6 & 56.9 & 41.3 & 41.2 & 73.3 & 69.4\\
SiamFC$++$~\cite{siamfcpp-aaai20}  & GoogleNet  & 303 & 15.8G & 54.4 & 56.9 & 38.6 & 36.9  & 75.4 & 70.5\\
Ocean~\cite{zhang2020ocean} $\Diamond$ & ResNet50  & 255 & 7.8G & 56.0 & 65.0 & 38.4 & 37.7 & 70.3 & 68.8\\
SiamBAN~\cite{siamban-cvpr20}  & ResNet50  & 255 & 12.1G & 51.4 & 52.1 & 41.0 & 41.7 & - & -\\
SiamAtt~\cite{siamrpnattn-cvpr20}   & ResNet50 & 255 & 7.8G & 56.0  & 64.8 & - & - &75.2 & -\\
TransT~\cite{transt-2021}  & ResNet50  & 256 & 29.3G & 64.9 & 73.8 & 50.7 & 51.7 & 81.4 & 80.3\\
TrDiMP~\cite{trsiam} $\Diamond$ & ResNet50  & 352 & 18.2G & 63.9 & - & - & -  &78.4 & 73.1\\
KeepTrack~\cite{keeptrack} $\Diamond$ & ResNet50  & 464 & 28.7G & 67.1 & 77.2 & - & - & - & -\\
AutoMatch~\cite{nascorr-corr21} & ResNet50  & - & - & 58.3 & - & 47.2 & 43.5 & 76.0 & 72.6\\
TransInMo$^*$~\cite{transinfo}  & ResNet50  & 255 & 16.9G & 65.7 & 76.0 & 52.0 & \blu{52.7} & - &-\\
STARK-S~\cite{stark}  & ResNet50  & 320 & 15.6G & 65.8 & - & - & - & 80.3 & -\\
STARK-ST~\cite{stark} $\Diamond$ & ResNet101  & 320 & 28.0G & 67.1 & 77.0 & - & - & \blu{82.0} & 
\gre{86.9}\\
\hline
\rowcolor{LightCyan} Sim-B$/$32  & ViT-B$/$32  & 320 & 11.5G & 66.2  & 76.1 & 51.1 & 48.1  & 79.1 & 83.9\\
\rowcolor{LightCyan} Sim-B$/$16$^*$  & ViT-B$/$16$^*$  & 224 & 14.7G & \blu{68.7}  & \blu{77.5} & \blu{53.7} & 52.6 & 81.5 & 86.0 \\
\rowcolor{LightCyan} Sim-B$/$16  & ViT-B$/$16  & 224 & 25.0G & \gre{69.3}  & \gre{78.5} & \gre{54.8} & \gre{53.8} & \gre{82.3} & \blu{86.5} \\
\rowcolor{LightCyan} Sim-L$/$14 & ViT-L$/$14  & 224 & 95.4G & \bai{70.5}  & \bai{79.7} & \bai{55.6} & \bai{55.7} & \bai{83.4} & \bai{87.4} \\
\Xhline{1pt} 
\end{tabular}}}
\end{center}
\end{table*}

We compare our SimTrack with other trackers on five datasets, including LaSOT~\cite{lasot-corr}, TNL2K~\cite{tnl2k-2021}, TrackingNet~\cite{trackingnet-eccv18}, UAV123~\cite{uav123-eccv16} and GOT-10k~\cite{GOT10k}.

{\flushleft {\bf{LaSOT}}} is a large-scale dataset with 1400 long videos in total. 
The $test$ set of LaSOT~\cite{lasot-corr} consists of 280 sequences. Table~\ref{table:sota} shows the AUC and normalized precision scores ($P_{norm}$) of all compared trackers. Our SimTrack can get a competitive or even better performance compared with state-of-the-art trackers. Our Sim-B$/16^*$ outperforms all compared trackers with a simpler framework and lower computation costs. Our Sim-B$/$16 achieves a new state-of-the-art result, 69.3\% AUC score and 78.5\% normalized precision score, with acceptable computation costs. After using the larger model ViT-L$/14$, our Sim-L$/$14 can get a much higher performance, 70.5\% AUC score and 79.7\% normalized precision score. We are the first to exploit such a large model and demonstrate its effectiveness in visual object tracking.

{\flushleft {\bf{TNL2K}}} is a recently published datasets which composes of 3000 sequences. We evaluate our SimTrack on the $test$ set with 700 videos. From Tab.~\ref{table:sota}, SimTrack performs the best among all compared trackers. The model with ViT-B$/$16 exceeds 2.8 AUC points than the highest AUC score (52.0\%) of all compared trackers. Leveraging a larger model can further improve the AUC score to 55.6\%.

{\flushleft {\bf{TrackingNet}}} is another large-scale dataset consists of 511 videos in the $test$ set. The $test$ dataset is not publicly available, so results should be submitted to an online server for performance evaluation. Compared with the other trackers with complicated interaction modules, our SimTrack is a more simple and generic framework, yet achieves competitive performance. By leveraging a larger model, Sim-L$/$14 outperforms all compared trackers including those with online update.

{\flushleft {\bf{UAV123 }}} provides 123 aerial videos captured from a UAV platform. In Table~\ref{tab:uav}, two versions of our method both achieve better AUC scores (69.8 and 71.2) than the highest AUC score (68.1) of all compared algorithms. 

{\flushleft {\bf{GOT-10k}}} requires training trackers with only the $train$ subset and testing models through an evaluation server. We follow this policy for all experiments on GOT-10k. As shown in Table~\ref{tab:got}, our tracker with ViT-B/16 obtains the best performance. When leveraging a larger model ViT-L/14, our model can further improve the performance to 69.8 AUC score.

\begin{table}[t]
\scalebox{0.90}{
\begin{tabular}{c?cccccccc}
\multirow{2}*{}    & \multirow{2}*{SiamFC} & \multirow{2}*{SiamRPN} & \multirow{2}*{SiamFC++} & \multirow{2}*{DiMP} & \multirow{2}*{TrDiMP} & \multirow{2}*{TransT} &  \multirow{2}*{\textbf{Ours}} & \multirow{2}*{\textbf{Ours}} \\ [3pt]
 & \cite{Bertinetto-ECCV16-SiamesFC} & \cite{siamrpn-cvpr18} & \cite{siamfcpp-aaai20} & \cite{dimp-iccv19} & \cite{trsiam} & \cite{transt-2021}  & ViT-B/16 & ViT-L/14 \\
\Xhline{1pt}
\multicolumn{1}{c?}{AUC$\uparrow$} & \multicolumn{1}{c}{48.5}   & \multicolumn{1}{c}{55.7}    & \multicolumn{1}{c}{63.1}   & 65.4  & \multicolumn{1}{c}{67.5}   & \multicolumn{1}{c}{\blu{68.1}}   &  \multicolumn{1}{c}{\gre{69.8}}        & \multicolumn{1}{c}{\bai{71.2}}        \\
\multicolumn{1}{c?}{Pre$\uparrow$}   & \multicolumn{1}{c}{64.8}   & \multicolumn{1}{c}{71.0}    & \multicolumn{1}{c}{76.9}   &85.6  & \multicolumn{1}{c}{87.2}   & \multicolumn{1}{c}{\blu{87.6}}   &  \multicolumn{1}{c}{\gre{89.6}}       & \multicolumn{1}{c}{\bai{91.6}}        \\
\end{tabular}}
\caption{Performance comparisons on UAV123~\cite{uav123-eccv16} dataset. \bai{Red}, \gre{green} and \blu{blue} fonts indicate the top-3 methods.}
\label{tab:uav}
\end{table}

\begin{table}[t]
\scalebox{0.90}{
\begin{tabular}{c?cccccccc}
\multirow{2}*{}    & \multirow{2}*{SiamFC} & \multirow{2}*{SiamRPN} & \multirow{2}*{SiamFC++} & \multirow{2}*{DiMP} & \multirow{2}*{TrDiMP} & \multirow{2}*{STARK-\tiny{S}} &  \multirow{2}*{\textbf{Ours}} & \multirow{2}*{\textbf{Ours}} \\ [3pt]
 & \cite{Bertinetto-ECCV16-SiamesFC} & \cite{siamrpn-cvpr18} & \cite{siamfcpp-aaai20} & \cite{dimp-iccv19} & \cite{trsiam} & \cite{stark}  & ViT-B/16 & ViT-L/14 \\
\Xhline{1pt}
\multicolumn{1}{c?}{AO$\uparrow$} & \multicolumn{1}{c}{34.8}   & \multicolumn{1}{c}{46.3}    & \multicolumn{1}{c}{59.5}  & 61.1   & \multicolumn{1}{c}{67.1}   & \multicolumn{1}{c}{\blu{67.2}}   &  \multicolumn{1}{c}{\gre{68.6}}        & \multicolumn{1}{c}{\bai{69.8}}        \\
\multicolumn{1}{c?}{$SR_{0.5}$ $\uparrow$}   & \multicolumn{1}{c}{35.3}   & \multicolumn{1}{c}{40.4}    & \multicolumn{1}{c}{69.5}  & 71.7  & \multicolumn{1}{c}{\blu{77.7}}   & \multicolumn{1}{c}{76.1}   &  \multicolumn{1}{c}{\gre{78.9}}        & \multicolumn{1}{c}{\bai{78.8}}        \\
\multicolumn{1}{c?}{$SR_{0.75}$ $\uparrow$}   & \multicolumn{1}{c}{9.8}   & \multicolumn{1}{c}{14.4}    & \multicolumn{1}{c}{47.9}  &  49.2 & \multicolumn{1}{c}{58.3}   & \multicolumn{1}{c}{\blu{61.2}}   &  \multicolumn{1}{c}{\gre{62.4}}        & \multicolumn{1}{c}{\bai{66.0}}        \\
\end{tabular}}
\caption{Experimental results on GOT-10k\_Test~\cite{GOT10k} dataset.}
\label{tab:got}
\end{table}

\subsection{Ablation Study and Analysis}
{\flushleft {\bf{Simplified Framework $vs.$ STARK-SV.}}}
To remove concerns about backbone, we compare our method with the baseline tracker STARK-SV~\cite{stark} using the same backbone architecture. In Table~\ref{tab:sim}, our design can consistently get significant performance gains with similar or even fewer computation costs. Our three variations with ViT-B$/32$, ViT-B$/16$ and ViT-L$/14$ as backbone outperforms STARK-SV for 3.7/3.1, 2.5/2.6 and 1.3/1.6 AUC points on LaSOT/TNL2K dataset, respectively, demonstrating the effectiveness and efficiency of our method.

{\flushleft {\bf{Training Loss $\&$ Accuracy.}}}
In Fig.~\ref{fig:loss}, we show the training losses and AUC scores of the baseline model STARK-SV and our method `Ours' on the LaSOT dataset. Both the two trackers utilize ViT-B$/$16 as the backbone. We can see that `Ours' uses fewer training epochs to get the same training loss with STARK-SV. When training models for the same epochs, `Ours' can get lower training losses than STARK-SV. In terms of testing accuracy, training our model for 200 epochs is enough to get the same AUC score (66.8\% $vs.$ 66.8\%) with the baseline model trained for 500 epochs. We think the main reason is `Ours' does not have a randomly initialized transformer head. The transformer head without pre-training needs more training epochs to get a good performance.

\begin{table}[t]
\begin{tabular}{c?ccccccc}
\multicolumn{1}{c?}{}      & \multicolumn{1}{c|}{\multirow{2}{*}{Backbone}} & \multicolumn{1}{c|}{\multirow{2}{*}{FLOPs}} & \multicolumn{3}{c|}{LaSOT}                                                                  & \multicolumn{2}{c}{TNL2K}                             \\ \cline{4-8} 
\multicolumn{1}{c?}{}      & \multicolumn{1}{c|}{}      & \multicolumn{1}{c|}{}                    & \multicolumn{1}{c|}{AUC$\uparrow$}       & \multicolumn{1}{c|}{P$_{norm}$$\uparrow$}   & \multicolumn{1}{c|}{P$\uparrow$}    & \multicolumn{1}{c|}{AUC$\uparrow$}  & \multicolumn{1}{c}{P$\uparrow$}    \\ \Xhline{1pt}
\multicolumn{1}{c?}{STARK-SV} & \multicolumn{1}{c|}{ViT-B/32}   & \multicolumn{1}{c|}{13.3G}               & \multicolumn{1}{l|}{62.5}      & \multicolumn{1}{l|}{72.1}      & \multicolumn{1}{l|}{64.0} & \multicolumn{1}{l|}{48.0} & \multicolumn{1}{l}{44.0} \\
\multicolumn{1}{c?}{Ours}  & \multicolumn{1}{c|}{ViT-B/32}    & \multicolumn{1}{c|}{11.5G}              & \multicolumn{1}{l|}{66.2 \tiny{\px{(+3.7)}}}      & \multicolumn{1}{l|}{76.1 \tiny{\px{(+4.0)}}}      & \multicolumn{1}{l|}{68.8 \tiny{\px{(+4.8)}}} & \multicolumn{1}{c|}{51.1 \tiny{\px{(+3.1)}}} & \multicolumn{1}{l}{48.1 \tiny{\px{(+4.1)}}} \\ \hline
\multicolumn{1}{c?}{STARK-SV} & \multicolumn{1}{c|}{ViT-B/16}    & \multicolumn{1}{c|}{25.6G}              & \multicolumn{1}{l|}{66.8}      & \multicolumn{1}{l|}{75.7}      & \multicolumn{1}{l|}{70.6} & \multicolumn{1}{l|}{52.2} & \multicolumn{1}{l}{51.1} \\
\multicolumn{1}{c?}{Ours}  & \multicolumn{1}{c|}{ViT-B/16}     & \multicolumn{1}{c|}{23.4G}             & \multicolumn{1}{l|}{69.3 \tiny{\px{(+2.5)}}} & \multicolumn{1}{l|}{78.5 \tiny{\px{(+2.8)}}} & \multicolumn{1}{l|}{74.0 \tiny{\px{(+3.4)}}} & \multicolumn{1}{l|}{54.8 \tiny{\px{(+2.6)}}} & \multicolumn{1}{l}{53.8 \tiny{\px{(+2.7)}}} \\ \hline
\multicolumn{1}{c?}{STARK-SV} & \multicolumn{1}{c|}{ViT-L/14}     & \multicolumn{1}{c|}{95.6G}             & \multicolumn{1}{l|}{69.2}          & \multicolumn{1}{l|}{78.2}          & \multicolumn{1}{l|}{74.3}     & \multicolumn{1}{l|}{54.0}     & \multicolumn{1}{l}{54.1}     \\
\multicolumn{1}{c?}{Ours}  & \multicolumn{1}{c|}{ViT-L/14}     & \multicolumn{1}{c|}{95.4G}             & \multicolumn{1}{l|}{70.5 \tiny{\px{(+1.3)}}}          & \multicolumn{1}{l|}{79.7 \tiny{\px{(+1.5)}}}          & \multicolumn{1}{l|}{76.2 \tiny{\px{(+1.9)}}}     & \multicolumn{1}{l|}{55.6 \tiny{\px{(+1.6)}}}     & \multicolumn{1}{l}{55.7 \tiny{\px{(+1.6)}}}     \\ 
\end{tabular}
\caption{Ablation study about our simplified framework and the baseline model STARK-S~\cite{stark}. `FLOPs' shows computation costs of different methods, AUC, P$_{norm}$ and P respectively denote AUC, normalized precision and precision.}
\label{tab:sim}
\end{table}

\begin{table}[t]
\RawFloats
	\begin{minipage}[c]{.52\textwidth}
		\centering
\begin{tabular}{|cl|l|l|l|l|l|}
\hline
\multicolumn{2}{|c|}{\#Num}                         & \ding{172}    & \ding{173}    & \ding{174}    & \ding{175}    & \ding{176}    \\ \hline
\multicolumn{2}{|c|}{Pretrain}                      & DeiT & Moco & SLIP & CLIP & MAE  \\ \hline
\multicolumn{1}{|c|}{\multirow{2}{*}{LaSOT}} & AUC  & 66.9 & 66.4 & 67.6 & 69.3 & 70.3 \\ \cline{2-7} 
\multicolumn{1}{|c|}{}                       & Prec & 70.3 & 69.4 & 71.0 & 74.0 & 75.5 \\ \hline
\multicolumn{1}{|c|}{\multirow{2}{*}{TNL2K}} & AUC  & 51.9 & 51.9 & 53.4 & 54.8 & 55.7 \\ \cline{2-7} 
\multicolumn{1}{|c|}{}                       & Prec & 49.6 & 49.4 & 51.8 & 53.8 & 55.8 \\ \hline
\end{tabular}
\caption{The AUC/Pre scores of SimTrack (with ViT-B$/16$ as backbone) when using different pre-training weights.}
\label{tab:pre}
	\end{minipage}
\hfill
	\begin{minipage}[c]{.45\textwidth}%
		\centering
    \includegraphics[width=1\linewidth]{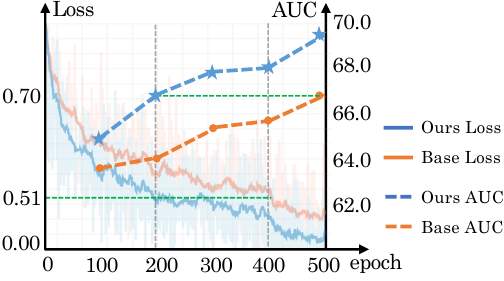}\\
    \caption{The training loss and AUC (on LaSOT) in the Y-axis for different training epochs (X-axis). 
    }
    \label{fig:loss}
	\end{minipage}
\end{table}

\noindent{\textbf{Results with Other Transformer Backbones.}}
We evaluate our framework with Swin Transformer~\cite{liu2021swin} and Pyramid Vision Transformer (PVT)~\cite{wang2021pvt}. 
For Swin Transformer, we made a necessary adaption, considering the shifted window strategy. We remove the $a(e^l, s^l)$ and $s^lW_V$ in the first function of Eq.(6), which has less influence according to our experiments (from 69.3\% to 69.1\% AUC score on LaSOT for SimTrack-ViT). 
The attention of each $search$ token is calculated with the tokens inside the local window and those from $exemplar$ features.
During attention calculation, the $exemplar$ features are pooled to the size of the local window.
For PVT, we reduce the reduction ratio of SRA module for the $exemplar$ by half, to keep a reasonable $exemplar$ size. 
In the Table~\ref{tab:swinpvt}, SimTrack with PVT-Medium is denoted as PVT-M and SimTrack with Swin-Base is denoted as Swin-B. PVT-M gets comparable AUC scores with fewer FLOPs, and Swin-B has higher AUC scores with similar FLOPs to STARK-S on both datasets, demonstrating the good generalization of our SimTrack.

\begin{table}[h]
\begin{tabular}{l|l|l|l|l|l|l}
\hline
     & DiMP & TrDiMP & TransT & STARK-S & {\ul \textbf{\textcolor{purple}{PVT-M}}} & {\ul \textbf{\textcolor{purple}{Swin-B}}} \\ \hline
FLOPs  & 5.4G & 18.2G   & 29.3G   & 15.6G    & {\ul \textbf{\textcolor{purple}{8.9G}}}  & {\ul \textbf{\textcolor{purple}{15.0G}}}   \\ \hline
LaSOT  & 56.9 & 63.9   & 64.9   & 65.8    & {\ul \textbf{\textcolor{purple}{66.6}}}  & {\ul \textbf{\textcolor{purple}{68.3}}}   \\ \hline
UAV123 & 65.4 & 67.5   & 68.1   & 68.2    & {\ul \textbf{\textcolor{purple}{68.5}}}  & {\ul \textbf{\textcolor{purple}{69.4}}}   \\ \hline
\end{tabular}
\caption{The AUC scores and FLOPs of SimTrack using PVT and Swin-Transformer as backbone on LaSOT and UAV123 dataset.}\label{tab:swinpvt}
\end{table}

{\flushleft {\bf{Different Pre-training.}}}
We evaluate our SimTrack when using ViT-B$/16$ as backbone and initializing the backbone with parameters pre-trained with several recent methods, including DeiT~\cite{deit2021}, MOCO-V3~\cite{mocov321}, SLIP~\cite{mu2021slip}, CLIP~\cite{clip21}, and MAE~\cite{mae}. From Table~\ref{tab:pre}, all of these versions achieve competitive performance with state-of-the-art trackers on the two datasets. However, the pre-trained parameters from MAE show the best performance, suggesting that appropriate parameter initialization is helpful to the training of SimTrack.

{\flushleft {\bf{Component-wise Analysis.}}}
To prove the efficiency of our method, we perform a component-wise analysis on the TNL2K~\cite{tnl2k-2021} benchmark, as shown in Table~\ref{tab:com}. The `Base' means STARK-SV with ViT-B$/16$, which obtains an AUC score of 52.2. In \ding{173}, `+Sim' indicates using our SimTrack framework without adding the distinguishable position embedding or foveal window strategy. It brings significant gains, $i.e.$ 1.3/1.4 point in terms of AUC/Pre score, and verifies the effectiveness of our framework. Adding our position embedding helps model performs slightly better (\ding{174} $vs.$ \ding{173}). Furthermore, the foveal window strategy brings an improvement of 0.8 point on AUC score in \ding{175}. This shows using more detailed target patches at the beginning contributes to improving accuracy.

\begin{figure*}[t]
    \centering
    \includegraphics[width=0.9\linewidth]{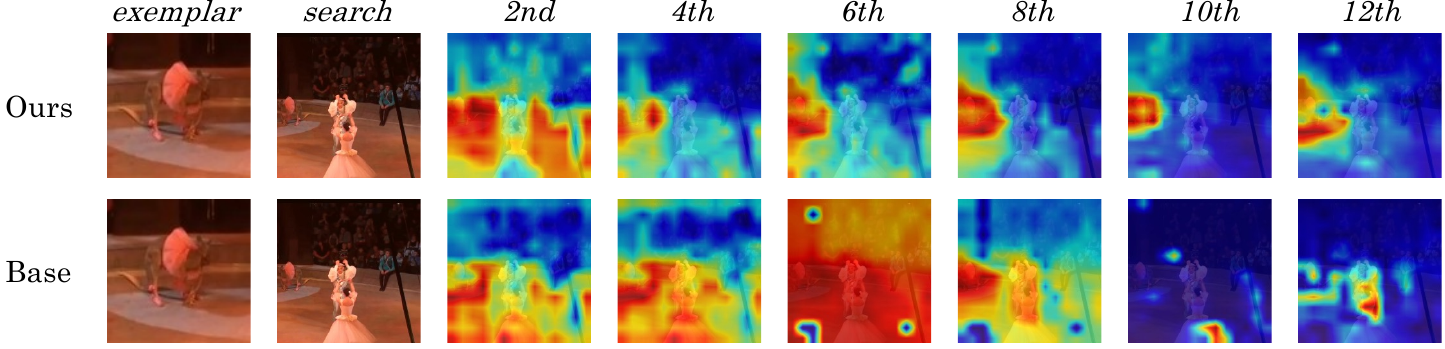}\\
    \caption{The images in different columns are the $exemplar$ image, $search$ image, target-relevant attention maps from the $2nd, 4th, 6th, 8th, 10th, 12th (last)$ layer of the transformer backbone. Details can be found in supplementary materials.}
    \label{fig:v1}
\end{figure*}

{\flushleft {\bf{Decoder Number.}}}
We analyze the necessity of introducing transformer decoders in our SimTrack. Specifically, we add a transformer decoder at the end of our backbone for further information interaction. In the decoder, the $search$ features from the backbone are used to get $query$ values. The $exemplar$ features are adopt to calculate $key$ and $value$. Through changing the layer number of the decoder from 0 to 6, the performance changes less. This shows another information interaction module is unnecessary in our framework, because our transformer backbone can provide enough information interaction between the $search$ and $exemplar$ features.

{\flushleft {\bf{Dense or Sparse Information Interaction.}}}
The information interaction between the $exemplar$ and $search$ features exist in all twelve blocks in our Sim-B/16, shown as \ding{172} in Table~\ref{tab:inter}.
In \ding{173}, we only enable the interaction in the $2nd, 4th, 6th, 8th, 10th$ and $12th$ block, removing half of the interaction in \ding{172}. As we can see, using less information interaction leads to 2.5 points AUC drop. When we further reduce half interaction in \ding{173}, the AUC score drops another 2.5 points in \ding{174}. The experiments show that comprehension information interaction helps to improve the tracking performance in SimTrack.

\begin{table}[t]
\RawFloats
  \begin{minipage}[]{.3\textwidth}
    \centering
        \begin{tabular}{c|l|c}
        \#Num &  Com & TNL2K$\uparrow$  \\ \hline
        \ding{172}     & Base   & 52.2/51.1 \\
        \ding{173}     & +Sim  & 53.5/52.5 \\
        \ding{174}     & +PosEm & 54.0/53.1 \\
        \ding{175}     & +FW & 54.8/53.8 \\
        \end{tabular}
        \caption{Component-wise analysis. AUC/Pre scores are reported respectively. The results demonstrate that each component is important in our framework. }\label{tab:com}
  \end{minipage}%
  \hfill
  \begin{minipage}[]{.3\textwidth}
   
    \centering
        \begin{tabular}{c|c|c}
        \#Num & Dec & TNL2K$\uparrow$       \\ \hline
        \ding{172}     & 0       & 54.8/53.8 \\
        \ding{173}     & 1       & 54.8/54.2 \\
        \ding{174}     & 3       & 54.6/54.0 \\
        \ding{175}     & 6       & 54.7/54.3 \\
        \end{tabular}
        \caption{The influence of introducing decoders in SimTrack. With sufficient interaction in the transformer backbone, decoder becomes redundant for SimTrack.}\label{tab:dec}
  \end{minipage}
  \hfill
  \begin{minipage}[]{.3\textwidth}
  
    \centering
        \begin{tabular}{c|c|c}
        \#Num & Ratio & TNL2K$\uparrow$       \\ \hline
        \ding{172}     & 100\%     & 54.8/53.8 \\
        \ding{173}     & 50\%       & 52.3/50.4 \\
        \ding{174}     & 25\%       & 49.8/46.2 \\
        \end{tabular}
        \caption{Analysis of information interaction ratio in backbone. \ding{172} is ours with interaction in 100\% blocks. \ding{173} and \ding{174} reduce the number of interaction blocks to 50\% and 25\%.}\label{tab:inter}
  \end{minipage}
\end{table}

{\flushleft {\bf{Visualization.}}}
Fig.~\ref{fig:v1} shows the target-relevant area in the search region for different layers. Our architecture can gradually and quickly focus on the designated target and keep following the target in the following layers. 
The visualization maps show that the Siamese backbone in `base' tends to learn general-object sensitive features instead of designated-target sensitive features and no information interaction hinders the backbone from `sensing' the target during feature learning. By contrast, `Ours' can produce designated-target sensitive features thanks to the information interaction from the first block to the last block.

\section{Conclusions}
This work presents SimTrack, a simple yet effective framework for visual object tracking. By leveraging a transformer backbone for joint feature learning and information interaction, our approach streamlines the tracking pipeline and eliminates most of the specialization in current tracking methods.
While it obtains compelling results against well-established baselines on five tracking benchmarks, both architecture and training techniques can be optimized for further performance improvements

{\flushleft {\bf{Acknowledgement.}}} This work was supported by the Australian Research Council Grant DP200103223, Australian Medical Research Future Fund MRFAI000085, CRC-P Smart Material Recovery Facility (SMRF) – Curby Soft Plastics, and CRC-P ARIA - Bionic Visual-Spatial Prosthesis for the Blind.

\bibliographystyle{splncs04}
\bibliography{egbib}

\newpage

\section{Appendix}
{\flushleft {\bf{Visualization.}}}
We show more target-relevant attention maps on the $search$ images in Fig.~\ref{fig:m0}. For `Base', we replace the backbone of STARK-S with ViT-B/16. `Ours' adopts the SimTrack framework with ViT-B/16 as the backbone. Both `Base' and `Ours' are trained with the same training setting as shown in the paper.
While target-relevant attention map can be obtained for `Ours' directly in the transformer backbone, `Base' does not have such information since $search$ and $exemplar$ are processed separately. To obtain the target-relevant attention map for `Base' model, we get the $exemplar$ and $search$ features from the $l_{th}$ transformer layer after training and calculate the $search$ attention weight $A(s^l)$ through (refer to Equ.(6) in the paper),
\begin{equation}
\small
\begin{gathered}
A(s^l) = softmax\left(\left[a(s^l, e^l), a(s^l, s^l)\right]\right),
\end{gathered}
\end{equation}
where $s^l\in\mathbb{R}^{{N_x}\times{D}}$, $e^l\in\mathbb{R}^{{N_z}\times{D}}$, $A(s^l)\in\mathbb{R}^{{N_x}\times{(N_z+N_x)}}$. We select the target-relevant part from $A(s^l)\in\mathbb{R}^{{N_x}\times{N_z}}$ and average it along the second dimension to get $A^*(s^l)\in\mathbb{R}^{{N_x}\times{1}}$. Then, we reshape $A^*(s^l)$ to $\frac{H_x}{s}\times\frac{W_x}{s}$ and up-sample it to the same size ($H_x\times{W_x})$ with the $search$ image. After that, we get the target-relevant attention maps as shown in Fig.~\ref{fig:m0}. 
As we can see in Fig.~\ref{fig:m0}, `Ours' can quickly and gradually focus on a more accurate and comprehensive target area because the vital information interaction in the backbone enables the $search$ feature learning to `sense' the designated target.

\begin{figure*}[h]
\vspace{-4mm}
    \includegraphics[width=1\linewidth]{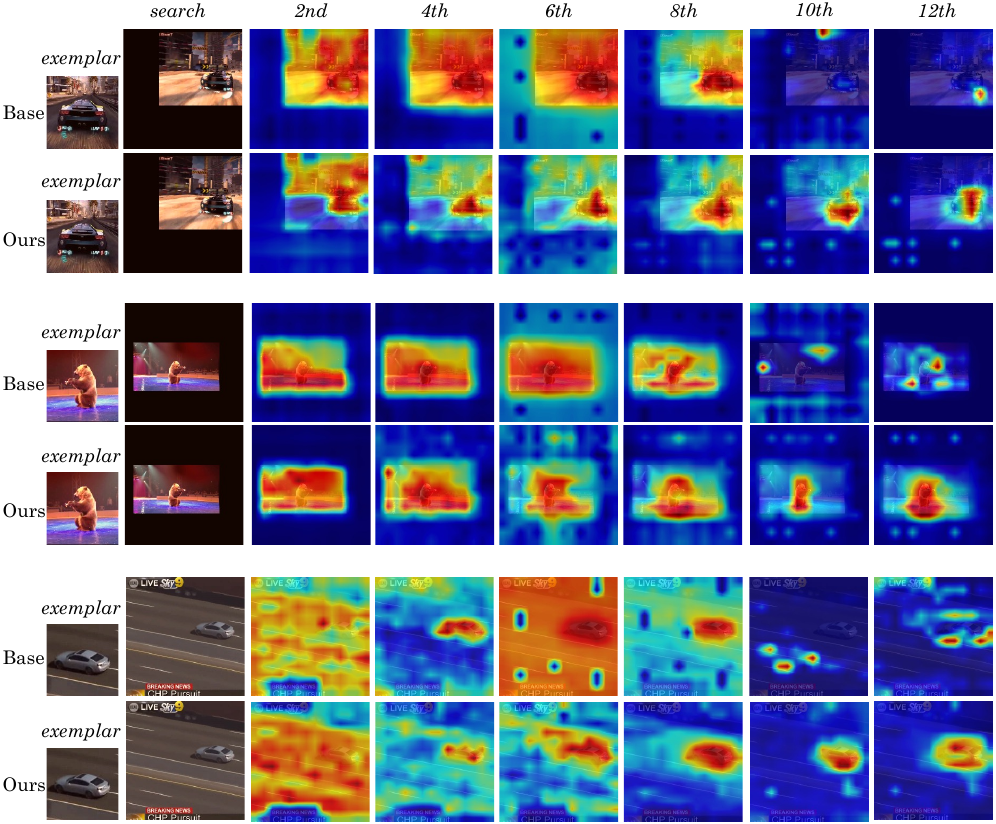}
    \caption{The images in different columns are the $exemplar$ image, the $search$ images, the target-relevant attention maps from the $2nd, 4th, 6th, 8th, 10th, 12th(last)$ layer of the transformer backbone. `Base' denotes the baseline model. `Ours' is our SimTrack. `Ours' can quickly and gradually focus on a more accurate and comprehensive target area.}
    \label{fig:m0}
\end{figure*}

{\flushleft {\bf{Training Details.}}}
The whole training needs 500 epochs with $6\times10^4$ image pairs in each epoch. The training batch size is 256.
All models are optimized with AdamW and the weight decay is $10^{-4}$. The initial learning rates of the backbone and head are $10^{-5}$ and $10^{-4}$, which will drop by a factor of 10 after 400 epochs. The loss weights $\lambda_{iou}$ and $\lambda_{L_1}$ are 2 and 5 in Equ.(3).
For Sim-B/32, we shift the $exemplar$ image by 16 pixels (half of the patch size 32) and crop a $64\times64$ foveal image in the centre of the shifted image.
For Sim-B/16, we directly crop a $64\times64$ foveal image in the centre of the $exemplar$ image. 
For Sim-L/14, to reduce computation cost, the input $exemplar$ size is reduced to $84\times84$. We centre crop a $42\times42$ image as the foveal image, where the partitioning lines are located in the centre of those on the $exemplar$ image. 

{\flushleft {\bf{Input Resolution.}}}
In the paper, we set the input size of $search$ image as $224\times224$ to be consistent with existing vision transformers.
We also evaluate the model performance when we increase the input resolution to $320\times{320}$ (the same with STARK-S) and $384\times{384}$. The results on LaSOT and TNL2K are shown in Table~\ref{tab:size}. A higher input resolution helps improve tracking accuracy. 

\begin{table}[t]
\begin{tabular}{c?l|ccc|cc}
\Xhline{1pt}
\multirow{2}{*}{\#Num} & \multirow{2}{*}{Input Size} & \multicolumn{3}{c|}{LaSOT}                                      & \multicolumn{2}{c}{TNL2K}           \\ \cline{3-7} 
                      &                           & \multicolumn{1}{c|}{AUC$\uparrow$}  & \multicolumn{1}{c|}{P$_{norm}$$\uparrow$} & P$\uparrow$    & \multicolumn{1}{c|}{AUC$\uparrow$}  & P$\uparrow$     \\ \Xhline{1pt}
\ding{172}                      & $224\times224$                    & \multicolumn{1}{c|}{69.3} & \multicolumn{1}{c|}{78.5}    & 74.0 & \multicolumn{1}{c|}{54.8} & 53.8  \\
\ding{173}                      & $320\times320$                   & \multicolumn{1}{c|}{70.0} & \multicolumn{1}{c|}{79.2}    & 74.8 & \multicolumn{1}{c|}{54.8} & 54.2  \\
\ding{174}                      & $384\times384$                      & \multicolumn{1}{c|}{70.4} & \multicolumn{1}{c|}{79.3}    & 75.0 & \multicolumn{1}{c|}{55.2} & 55.2  \\
\Xhline{1pt}
\end{tabular}
\caption{The performance of SimTrack (with ViT-B$/16$ as backbone) with diverse input sizes. A higher input resolution helps improve tracking accuracy. }
\label{tab:size}
\end{table}

\end{document}